**Manuscript Title: Human-in-the-Loop Large Language Model Framework for Identification of Cutaneous Immune-Related Adverse Events**

Charles Lu[1], Olivia Burke[1], Debby Cheng[1], Adam Kashlan[1], Caitlyn Duffy[1], Zeyun Lu[1], Lirit Fuksman[1], Jin Ning Tian[1], Andrew Sedlack[1], Priya Katyal[1], Eudora Lee[1], Ralina Karagenova[1], Chuck Lin[1], Kun-Hsing Yu[3,6,7], Nicole LeBoeuf[1], Alexander Gusev[1], Yevgeniy R. Semenov[1,2,3,4,5]**

**Corresponding author, Email: ysemenov@mgh.harvard.edu, Twitter handle: @EugeneSemenovMD

[1]Department of Dermatology, Massachusetts General Hospital, Boston, MA, USA.

[2]Harvard Medical School, Boston, MA, USA.

[3]Harvard Data Sciences Initiative, Boston, MA, USA.

[4]Harvard Program in Therapeutic Sciences, Boston, MA, USA.

[5]Cancer Data Sciences Program, Dana-Farber/Harvard Cancer Center, Boston, MA, USA.

[6]Department of Biomedical Informatics, Harvard Medical School, Boston, MA, USA.

[7]Department of Pathology, Brigham and Women's Hospital, Boston, MA, USA.

**Keywords:** cutaneous immune-related adverse events; immune checkpoint inhibitors; large language models; retrieval-augmented generation; human-in-the-loop; electronic health records; adverse event detection; clinical note abstraction; oncology; augmented intelligence

**Declarations**

**Consent for publication**: Not applicable

**Corresponding Author**: Yevgeniy R. Semenov, MD, MA, ysemenov@mgh.harvard.edu

**Competing interests:** Not applicable

**Authors' contributions:** C.L. and O.B. contributed to study conception, data collection, data curation, and drafting of the manuscript. D.C., A.K., C.D., and Z.L. assisted with data acquisition and management. L.F., J.N.T., A.S., P.K., E.L., C.L.-2,and R.K. contributed to drafting and revising the manuscript. K.-H.Y. and A.G. provided analytical support and contributed to data interpretation. N.L. offered clinical expertise and guidance on contextualizing findings. Y.R.S. conceived and supervised the study, oversaw project administration, provided critical revisions, and is the guarantor of the work. All authors reviewed and approved the final manuscript.

**Acknowledgements:** None

**Abstract:**

This study evaluated a retrieval-augmented, multi-agent large language model (LLM)–driven, human-in-the-loop framework for detecting cutaneous immune-related adverse events (cirAEs) from clinical notes. Compared with unassisted manual review, the LLM-assisted workflow improved accuracy (F1 = 0.88 vs 0.77), inter-rater agreement measured by Cohen's $\kappa$ ($\kappa = 0.82$ vs 0.50), and reduced average review time by approximately half. This framework pilots how LLMs can be applied to identify immune-related toxicities across organ systems and, more broadly, enable accurate, scalable, and transparent adverse event data extraction.

**Main text:**

Immune-related adverse events (irAEs) have emerged as both clinically meaningful markers of immune activation and important constraints on the safe delivery of immune checkpoint inhibitors (ICIs). Cutaneous irAEs (cirAEs) are among the earliest and most frequently encountered toxic effects, sometimes preceding noncutaneous irAEs and, in some studies, associated with improved oncologic outcomes.[1] Their identification has practical consequences: early recognition can expedite toxicity management without unnecessarily compromising ICI therapy. However, determining whether a dermatologic finding truly represents a cirAE remains difficult in routine practice and retrospective study. Manual review of unstructured electronic health record text is time-consuming, variable across reviewers, and difficult to scale. This problem is especially pronounced for cirAEs, which span diverse morphologic patterns, have broad clinical differentials, and often require dermatology-specific interpretation of context, timing, and competing causes. As a result, cirAEs provide a demanding test case for automated or semi-automated approaches to clinical phenotyping.

Natural language processing and machine learning approaches for irAE detection have shown promise, but performance has been variable and clinical integration remains limited. Gupta et al. used keyword filtering followed by deep learning classifiers, including convolutional neural networks and bidirectional long short-term memory models, achieving an F1 score of 0.75.[2] Barman et al. fine-tuned SciBERT for toxicity-specific classification and reported accuracies approaching 85%.[3] Yet both studies treated irAE ascertainment as a fully automated prediction problem, not as a collaborative workflow that supports expert review. Recent studies have turned to large language models. Guillot et al. applied GPT-4 to CAR-T progress notes and reported 64% accuracy,[4] whereas Sun et al. used retrieval-augmented Mistral 7B OpenOrca to detect myocarditis with 95% sensitivity and specificity.[5] irAE-GPT extended similar methods across electronic health record and clinical trial data.[6] Still, these frameworks were developed primarily for automated detection in comparatively standardized documentation environments, often inpatient. Whether they generalize to outpatient cirAE adjudication is uncertain, given the greater diagnostic ambiguity, broader differential diagnosis, and reliance on dermatology-specific clinical interpretation.

To address this gap, we developed a retrieval-augmented, multi-agent large language model framework for human-in-the-loop review of suspected irAEs. The system combines automated retrieval of relevant clinical text with discrete LLM-driven reasoning roles within a coordinated workflow, or agents, where separate prompts guide the model to perform distinct analytic tasks: a primary “synthesizer” agent compiles reasoning traces, and a secondary “critic” agent validates internal consistency. This design is meant to assist, not replace, expert adjudication by consolidating fragmented documentation and making the basis for each conclusion available for reviewer verification. We applied the framework to cirAEs, a stringent use case because dermatologic toxic effects in outpatient practice are morphologically diverse, diagnostically ambiguous, and often documented across multiple encounters. This framework not only addresses a key informatics barrier but also targets a clinically critical problem whose resolution directly impacts patient safety and real-world immunotherapy monitoring.

To assess the framework's performance and generalizability, we benchmarked leading open- and closed-weight models (LLaMa-3.1-70B, Gemma-2-27B, and Qwen-2.5-72B) in a zero-shot setting using a multi-institutional dataset of annotated EHR notes from Massachusetts General Hospital, Brigham and Women's Hospital, and Dana-Farber Cancer Institute. On a holdout subset of 100 patients' clinical notes, the GPT-4o agentic pipeline achieved the highest performance, with a sensitivity of 0.82 and specificity of 0.88, compared with LLaMa-3.1 (0.55, 0.63), Gemma-2 (0.68, 0.64), and Qwen-2.5 (0.23, 0.94), and was used as the reasoning core in the final pipeline (Fig. 2).

In a comparative study, two groups of reviewers assessed patient records: one using unassisted manual review (group A) and the other using the LLM-assisted pipeline (group B). Ground-truth labels were established by a board-certified oncodermatologist. Across all evaluated performance metrics, group B exceeded group A. Sensitivity, specificity, precision, and F1 score were 0.91, 0.92, 0.86, and 0.88 in the LLM-assisted group, compared with 0.79, 0.87, 0.76, and 0.77 in the manual-review group (Fig. 3A). Interrater reliability likewise increased, with κ improving from 0.50 in group A to 0.82 in group B ($p < 0.001$), and raw agreement increasing from 76.5% to 92.6% (Fig. 3B). Poststudy survey responses suggested that the augmented workflow reduced average review time by approximately 50% and was rated helpful or very helpful in 70% of cases (Fig. 3C-D).

These findings suggest that structured LLM augmentation can improve the accuracy, consistency, and efficiency of clinical text review. Notably, reviewers using the LLM-assisted workflow outperformed the model operating autonomously, supporting human-in-the-loop collaboration for and supporting the view that LLMs are most effective when used to scaffold rather than supplant clinical expertise. In the context of cirAE detection, improved agreement and reduced review burden may facilitate more timely and reliable monitoring of immunotherapy toxicities in real-world practice, an area of clear relevance to both oncodermatology and immuno-oncology.

Our framework directly addresses prior limitations by operating entirely in zero-shot, human-in-the-loop design that integrates expert-guided retrieval-augmented reasoning to provide interpretable, scalable clinical abstraction without task-specific fine-tuning. Grounded in retrieval-augmented reasoning informed by domain expertise, it achieves high performance without additional tuning and remains deployable across institutions with heterogeneous documentation practices. Its modular, agentic architecture is readily extensible to other irAEs and safety endpoints, providing transparent reasoning traces that enhance reviewer confidence and support scalable, reproducible clinical data curation.

As EHR data volume and complexity continue to grow, the bottleneck in real-world research increasingly lies in reliable data curation. Human-in-the-loop LLM systems offer a practical balance between automation and accountability, combining scalable contextual reasoning with expert oversight. The observed 50% reduction in review time suggests that such workflows could substantially accelerate dataset generation for immuno-oncology and other data-intensive clinical domains.

In sum, our retrieval-augmented, agentic LLM framework substantially improved the accuracy, concordance, and efficiency of cirAE identification from unstructured clinical notes both in the inpatient and outpatient clinical setting. By coupling transparent reasoning with expert oversight, it demonstrates how LLMs can enhance reproducibility and scalability in clinical data abstraction. Although evaluated in dermatologic toxicity, the system's modular design enables broad application to other irAEs and complex phenotyping tasks, illustrating how agentic LLMs can augment human expertise to enhance precision, reproducibility, and scalability in real-world evidence generation—ultimately advancing safer, more responsive care.

## Methods

### *Ethics Approval*

The Mass General Brigham Institutional Review Board approved the study (Protocol #2020P002307). The study meets the criteria of secondary research, for which consent is not required. The research involving human participants was conducted in accordance with the Declaration of Helsinki. The AI-specific ethical aspects followed the UNESCO Recommendation on the Ethics of Artificial Intelligence (2021) and the Ethics Guidelines for Trustworthy AI (European Commission, 2019)

### *Study Design and Cohort*

This retrospective study compared cirAE ascertainment by human reviewers, the current standard approach, with and without LLM assistance. We randomly selected 160 patients from a clinical database of individuals treated with ICIs (Table 1). For each patient, unstructured clinical notes were obtained from the electronic health record from 2 weeks before treatment initiation through 2 years and 90 days after the first ICI dose. All data were processed within a secure institutional computing environment in accordance with institutional governance policies, with record access restricted to authorized study personnel. The 90-day extension was intended to capture events documented shortly after the end of the prespecified 2-year follow-up interval.

### *Ground Truth Establishment*

The reference standard for the presence or absence of cirAEs for each patient was established through manual chart review by a board-certified dermatologist specializing in supportive oncodermatology. This expert determination served as the ground truth against which all other reviews were compared.

*Patient Note preprocessing.*

Clinical notes underwent preprocessing to remove duplicated or boilerplate text, followed by segmentation into semantic chunks via mxbai-embed-large[7] and indexing into a vector database utilizing ChromaDB.[8] For each patient, a hybrid retrieval system combining dense and sparse vector search gathered a comprehensive context from the indexed notes.

*LLM benchmarking and Prompt Engineering*

A random sample of 100 previously manually phenotyped, preprocessed clinical notes was used to evaluate the zero-shot performance of three state-of-the-art open-source LLMs (LLaMa-3.1, Gemma-2, and Qwen-2.5) alongside GPT-4o. Each model received the same prompt, and outputs were assessed against the established ground truth. Prompts were iteratively refined based on average model performance by an analyst with input from a domain-expert (Supplementary Methods). 10 iterations of prompt engineering were undertaken, with an iteration defined as a prompt which increases F1 score performance for all models (Supplemental methods 1). The best performing LLM on the final iteration of the prompt was used to create the agents in the full cirAE identification pipeline.

*LLM-Assisted Review Pipeline*

An agentic pipeline was implemented within a retrieval-augmented generation framework (Figure 1), in which separate LLM agents, defined by role-specific prompts, performed sequential synthesis and critique over retrieved clinical evidence. A handcrafted prompt was first applied to a primary LLM, and relevant information was retrieved from a vector database using a hybrid search strategy that combined dense semantic retrieval with sparse keyword-based BM25S search. Keywords were selected with domain expert input to capture both common descriptions of cirAEs and frequent negation patterns. The retrieved passages were then synthesized by the primary agent to generate an initial binary classification, which was subsequently evaluated by a secondary critic agent as a quality-control step. The output shown to human reviewers included both the binary recommendation and the specific note excerpts supporting that conclusion. Ablation study was performed to demonstrate the utility of the critique agent (Supplemental Figure 1). Model inference was performed using a HIPAA-compliant deployment of GPT-4o hosted through Microsoft Azure OpenAI Services within the Mass General Brigham secure computing environment. Clinical notes were processed entirely within this infrastructure and were not transmitted to public model endpoints.

*Reviewer Groups and Process*

Two pairs of nondermatologist reviewers, each of whom received structured training in cirAE adjudication, were blinded to the ground-truth labels and tasked with determining whether `` in the study cohort had developed cirAEs. Training was provided by a board-certified oncodermatologist and included instruction on the clinical presentation and terminology of common cirAEs, as well as practice review of representative cases before study initiation. The reviewers were divided into 2 groups. Group A performed standard manual review of the clinical notes without algorithmic assistance. Group B used the human-in-the-loop workflow, in which the LLM pipeline generated an initial assessment with supporting evidence that the reviewer could verify, modify, or accept before making a final determination.

*Statistical Analysis*

The primary outcomes were review accuracy and inter-rater agreement, secondary outcomes included subjective 'helpfulness' grading of the LLM pipeline and self-reported average time per chart recorded by reviewers in the post-study survey. Accuracy was measured against the oncodermatologist's ground truth using sensitivity, specificity, precision, and the F1 score. Inter-rater agreement within each group was quantified using Cohen's $\kappa$ and raw percentage agreement. A Wilcoxon signed-rank test was used to determine the statistical significance of the difference in Cohen's $\kappa$ values between groups. Reviewer subjective 'helpfulness' grading and self-reported time per chart were assessed via a post-study participant survey. P values of 0.05 or less were regarded as significant. All analysts were blinded to the ground truth.

*Ethics declarations:*

Funding: Y.R.S. is an advisory board member/consultant and has received honoraria from Arcutis Inc, Alterome Therapeutics, Castle Biosciences, Galderma, Incyte Corporation, Iovance Biotherapeutics, Pfizer Inc, Regeneron, Sanofi. All of these activities are not related to this work. K.-H.Y. is supported in part by the National Institute of General Medical Sciences grant R35GM142879, the National Heart, Lung, and Blood Institute grant R01HL174679, the Department of Defense Peer Reviewed Cancer Research Program Career Development Award HT9425-231-0523, and the Research Scholar Grant RSG-24-1253761-01-ESED (grant DOI: https://doi.org/10.53354/ACS.RSG-24-1253761-01-ESED.pc.gr.193749) from the American Cancer Society.

*Data Availability:*

The datasets generated and/or analyzed during the current study are available from the corresponding author upon reasonable request.

**Figure Legends:**

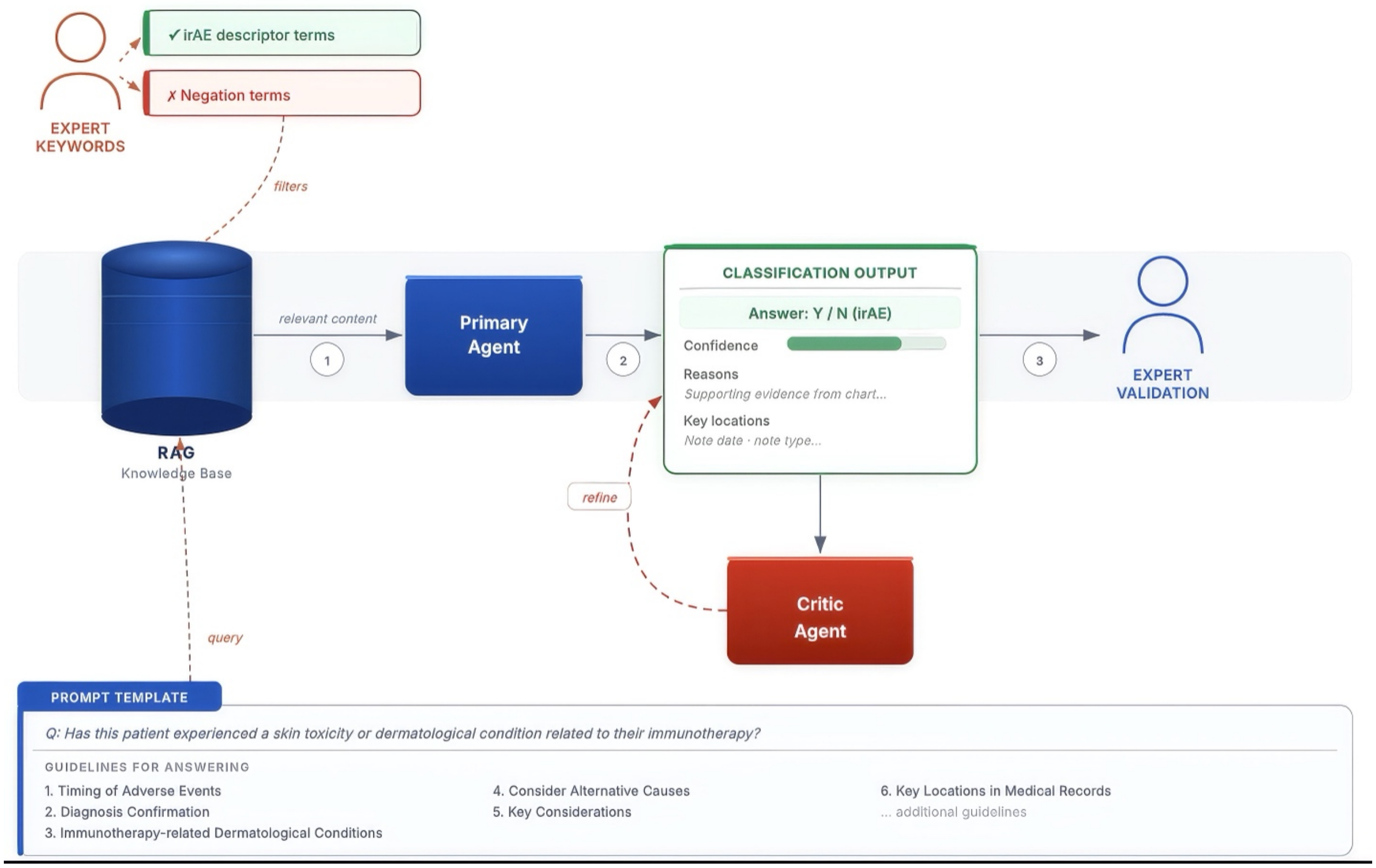

**Figure 1: Retrieval-Augmented Generation (RAG) Pipeline for cirAE Detection.** The diagram illustrates the sequential workflow of the agentic pipeline. The process begins with a hybrid search and re-ranking of text from a vector database, followed by synthesis and critique by two distinct LLM agents, and concludes with the presentation of a binary assessment and supporting evidence to the human reviewer. An example of the engineered prompt used to guide the primary LLM agent is also shown.

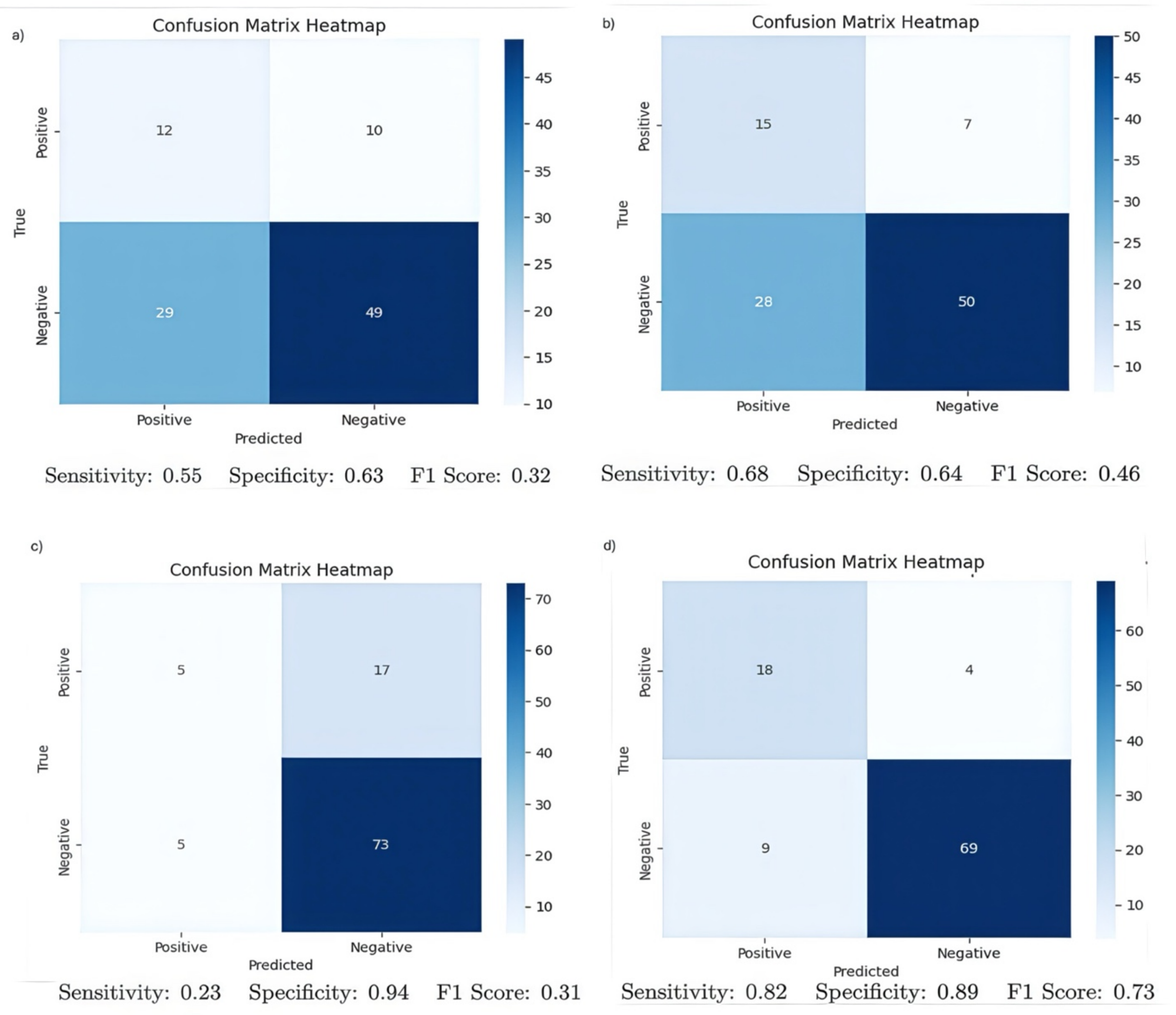

**Figure 2. Confusion matrices obtained with identical zero-temperature prompts.** (a) LLaMA-3.1-70B-Instruct-Q5_K_M, (b) Gemma-2-27B-it-Q5_K_M, (c) Qwen2.5-72B-Instruct-Q5_K_M via Ollama 0.3.12, and (d) GPT-4o (2024-08-06) via Azure OpenAI (full precision). All evaluations used the same Lambda workstation (dual 32-core AMD EPYC 7543 3.70 GHz, 128 GB DDR4-3200 RAM, 3.84 TB NVMe SSD, 2× NVIDIA Titan RTX 24 GB with NVLink).

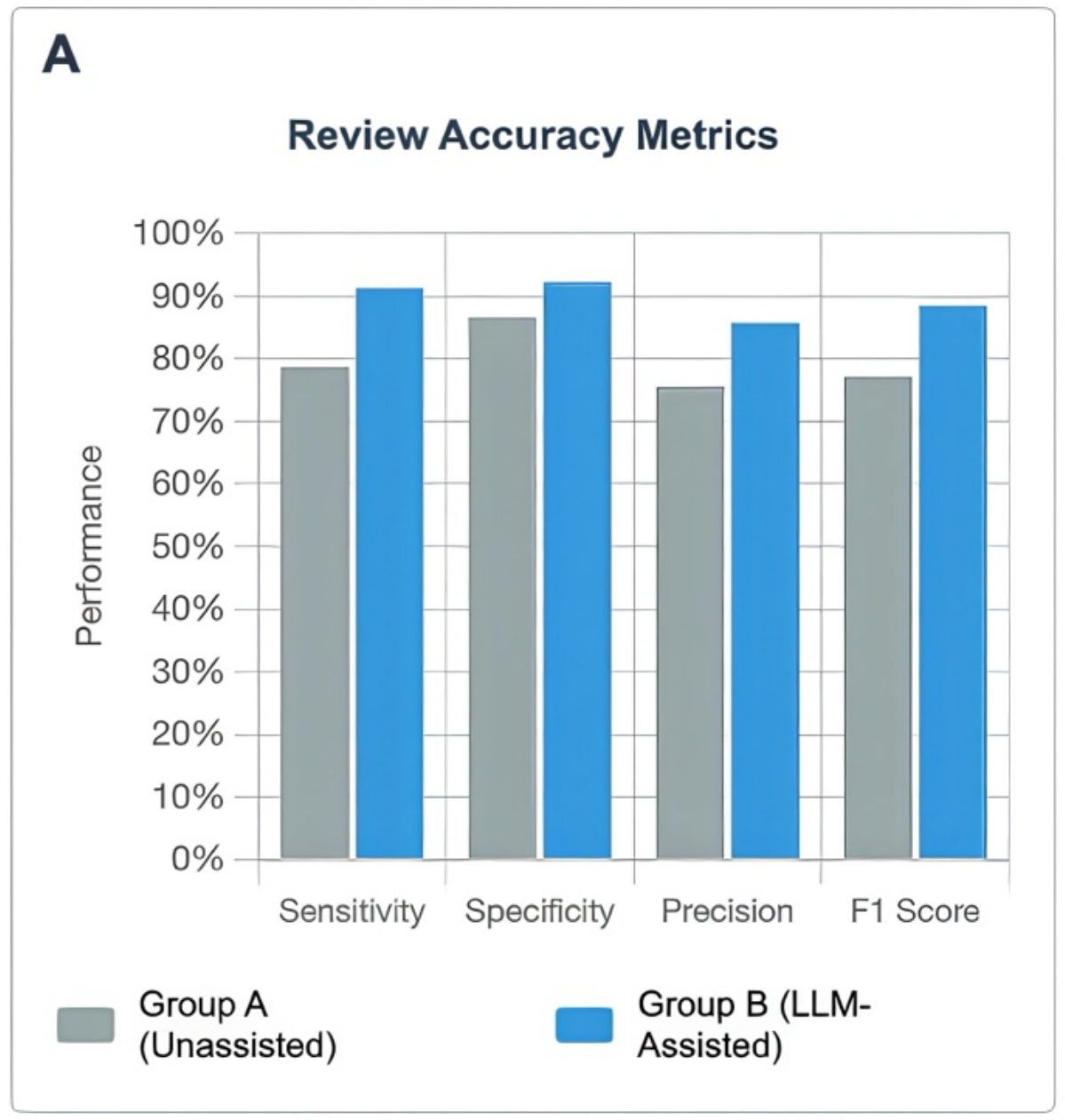

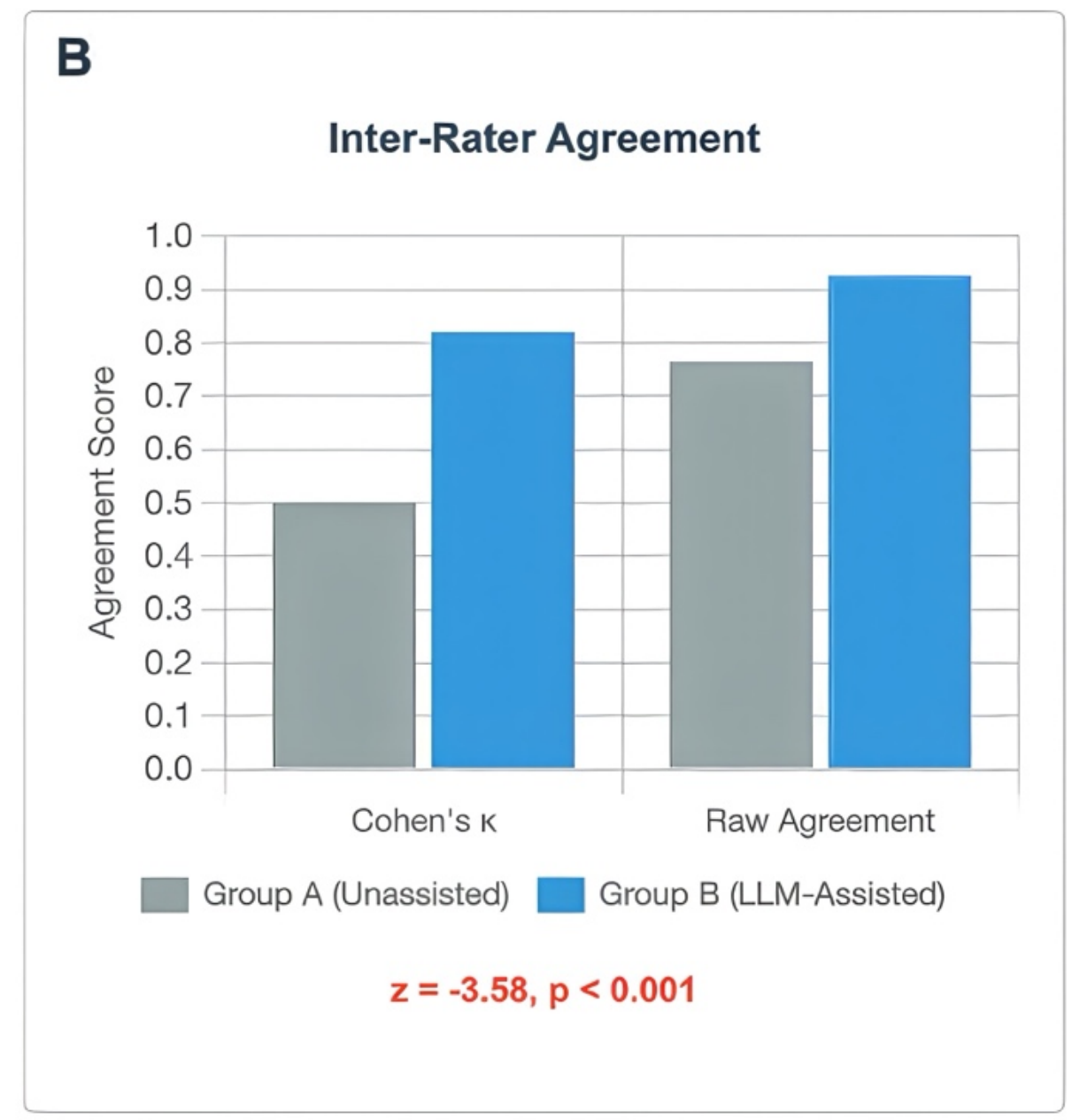

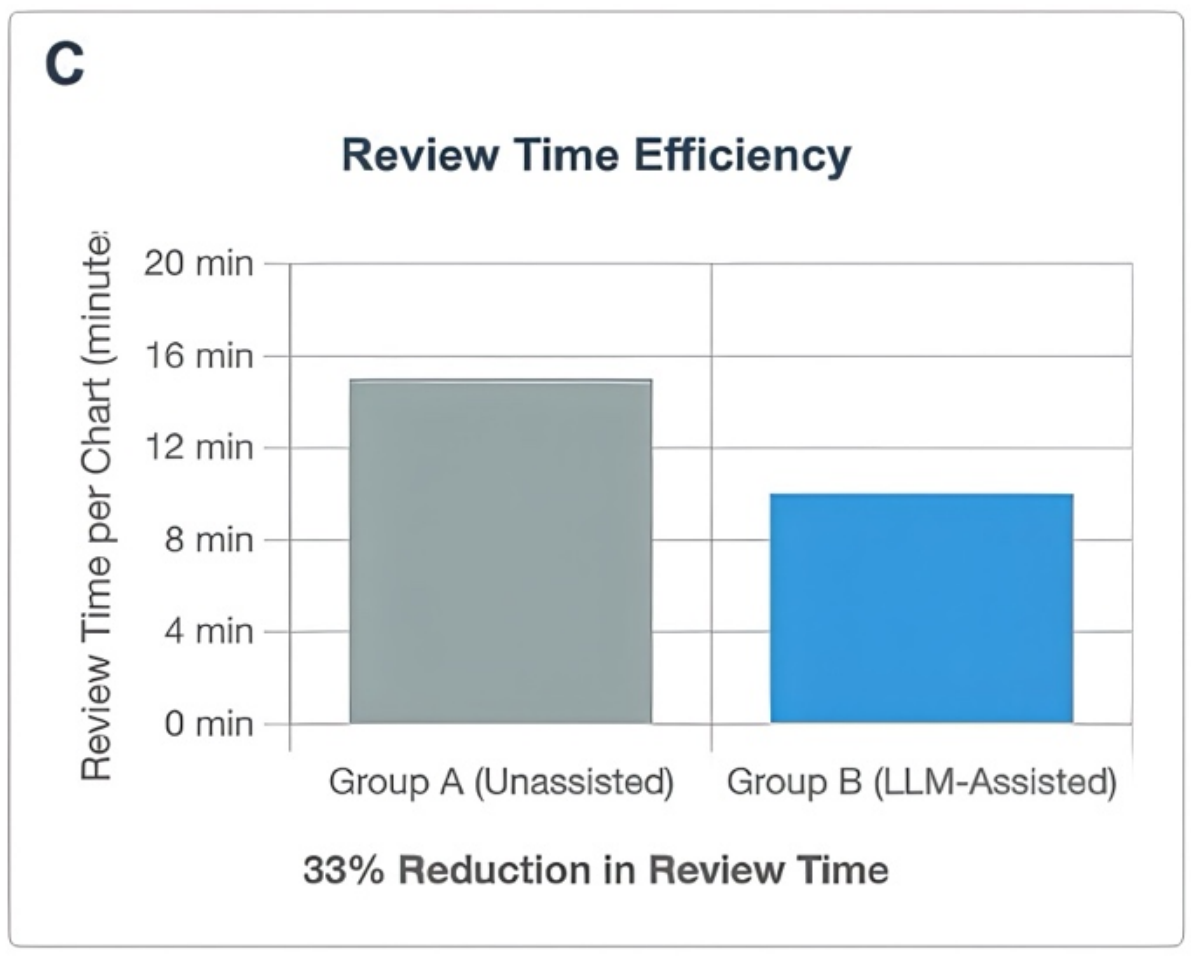

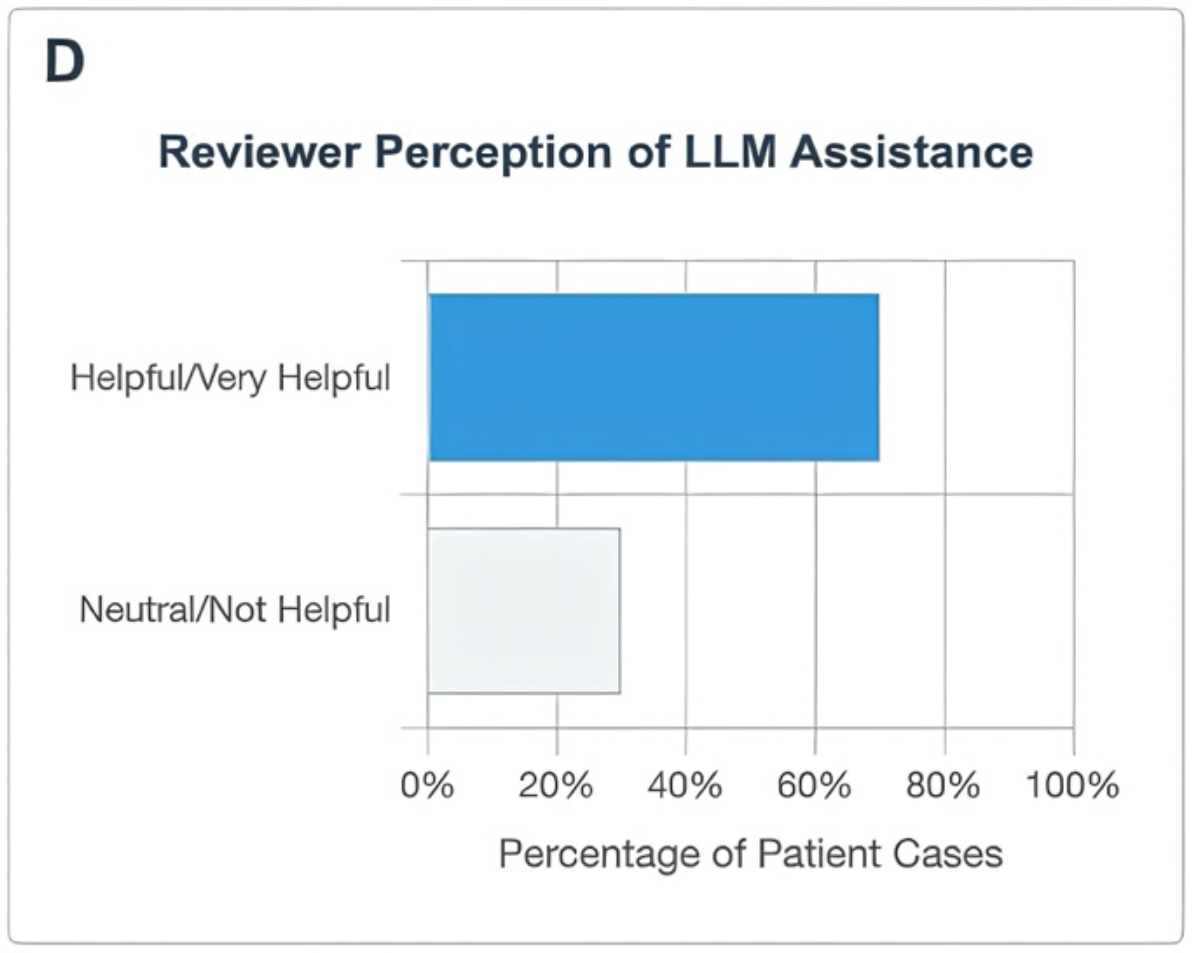

**Figure 3: HIL Comparative Summary** (a) Reviewer accuracy as measured against ground truth defined by a board-certified dermatologist. (b) Inter-rater agreement measured by Cohen's κ. (c) Review time efficiency increased as reported by the rater survey (15 vs 10 minutes). (d) Reviewer perception of LLM effectiveness over all cases.